\documentclass[review]{elsarticle}

\usepackage{lineno,hyperref}
\modulolinenumbers[5]

\usepackage{amsmath}
\usepackage{subfigure}
\usepackage{fixltx2e}
\usepackage{dblfloatfix}
\usepackage{array}
\usepackage{mdwmath}
\usepackage{mdwtab}
\usepackage{fixltx2e}
\usepackage{url}
\usepackage{color}
\usepackage[]{graphicx}
\usepackage{epsfig}
\usepackage{bbm}
\usepackage{enumerate}
\usepackage{amsfonts}
\usepackage{amsmath}
\usepackage{amssymb}
\usepackage{algorithm}
\usepackage{algpseudocode}
\usepackage{eqparbox}
\usepackage{comment}

\journal{Ultramicroscopy}
\begin{document}

\begin{frontmatter}
\title{Reduced Electron Exposure for Energy-Dispersive Spectroscopy using Dynamic Sampling}

\author{Yan Zhang \fnref{myfootnote}}
\author{G. M. Dilshan Godaliyadda\fnref{myfootnote,myfootnote1}}
\author{Nicola Ferrier\fnref{myfootnote2}}
\author{Emine B. Gulsoy\fnref{myfootnote3}}
\author{Charles A. Bouman\fnref{myfootnote1}}
\author{Charudatta Phatak\corref{correspAuth}\fnref{myfootnote}}
\cortext[correspAuth]{Corresponding author}
\ead{cd@anl.gov}

\fntext[myfootnote]{Materials Science Division, Argonne National Laboratory, 9700 Cass Ave, Lemont, IL 60439}
\fntext[myfootnote1]{ECE Department, Purdue University, 465 Northwestern Ave, West Lafayette, IN 47907}
\fntext[myfootnote2]{Mathematics and Computer Science Division, Argonne National Laboratory, 9700 Cass Ave, Lemont, IL 60439} 
\fntext[myfootnote3]{Department of Materials Science and Engineering, Northwestern University, 2220 Campus Drive, Evanston, IL 60208}

\begin{abstract}

Analytical electron microscopy and spectroscopy of biological specimens, polymers, 
and other beam sensitive materials  has been a challenging area due to irradiation damage. There is a pressing need to develop novel imaging and spectroscopic imaging methods that will minimize such sample damage as well as reduce the data acquisition time.
The latter is useful for high-throughput analysis of materials structure and chemistry. In this work, we present a novel machine learning based method for dynamic sparse sampling of EDS data using a scanning electron microscope. Our method, based on the supervised learning approach for dynamic sampling algorithm and neural networks based classification of EDS data, allows a dramatic reduction in the total sampling of up to 90\%, while maintaining the fidelity of the reconstructed elemental maps and spectroscopic data.  We believe this approach will enable imaging and elemental mapping of materials that would otherwise be inaccessible to these analysis techniques. 
\end{abstract}

\begin{keyword}
scanning electron microscopy (SEM) \sep Energy dispersive spectroscopy (EDS) \sep dynamic sampling \sep SLADS \sep Neural Networks \sep dose reduction
\end{keyword}

\end{frontmatter}


\section{Introduction}
Analytical electron microscopy based on energy dispersive X-ray spectroscopy (EDS) is a very versatile and successful technique for exploring elemental composition in microanalysis from the sub-nanometer scale to the micron scale \cite{goldstein2012scanning,Allen2010,Krivanek2012}.  Modern  scanning electron microscopes (SEM) equipped with EDS detectors are routinely used for qualitative, semi-quantitative or quantitative elemental mapping  of various materials ranging from inorganic to organic, and including biological specimens.  
Although EDS allows us to identify the elemental composition at a given location with high accuracy, each spot measurement can take anywhere from 0.1-10 s to acquire. As a result, if one wants to acquire EDS maps on a rectilinear grid with $256 \times 256$ grid points, the total imaging time could be on the order of tens to hundreds of hours. Furthermore, during the acquisition process, the sample gets exposed to a highly focused electron beam that can result in unwanted radiation damage such as knock-on damage, radiolysis, sample charging or heating. Organic and biological specimens are more prone to such damage due to electrostatic charging. Therefore minimizing the total radiation exposure of the sample is also of critical importance. One approach to solve this problem is to sample the rectilinear grid sparsely. However, it is critical that elemental composition maps reconstructed from these samples are accurate. Hence the selection of the measurement locations is of critical importance. 

Sparse sampling techniques in the literature fall into two main categories -- Static Sampling and Dynamic Sampling (DS). In Static Sampling the measurement locations are predetermined. Such methods include object independent static sampling methods such as Random Sampling strategies \cite{Hyrum13} and Low-discrepancy Sampling strategies \cite{LDSampling}, and sampling methods based on a model of the object being sampled such as those described in \cite{Mueller2011,wang2010variable}. In Dynamic Sampling, previous measurements are used to determine the next measurement or measurements. Hence, DS methods have the potential to find a sparse set of measurements that will allow for a high-fidelity reconstruction of the underlying sample. DS methods in the literature include dynamic compressive sensing methods \cite{seeger2008compressed,carson2012communications} which are meant for unconstrained measurements, application specific DS methods \cite{Seeger10,joost2012dynamic,Vanlier2012}, and point-wise DS methods \cite{merryman2005adaptive,godaliyaddaMBDS,Godaliyadda2}. In this paper, we use the dynamic sampling method described in \cite{Godaliyadda2}, Supervised Learning Approach for Dynamic Sampling (SLADS). SLADS is designed for point-wise measurement schemes, and is both fast and accurate, making it an ideal candidate for EDS mapping. 

In SLADS, each measurement is assumed to be scalar valued, but each EDS measurement, or spectrum, is a vector, containing the electron counts for different energies. Therefore, in order to apply SLADS for EDS, we need to extend SLADS to vector quantities or convert the EDS spectra into scalar values. In particular, we need to classify every measured spectrum as pure noise or as one of $L$ different phases. To determine whether a spectrum is pure noise, we use a Neural Network Regression (NNR) Model \cite{specht1991general}. For the classification step we use Convolutional Neural Networks (CNNs).

Classification is a classical and popular machine learning problem in computer science for which many well-established models and algorithms are available. Examples include logistic regression and Support Vector Machines (SVM) which have been proven very accurate for binary classification \cite{suykens1999least}. Artificial neural networks, previously known as multilayer perceptron, have recently gained popularity for multi-class classification particularly because of CNNs \cite{hinton2006reducing,lecun2015deep} that introduced the concept of deep learning. The CNNs architecture has convolution layers and sub-sampling layers that extract features from input data before they reach fully connected layers, which are identical to traditional neural networks. CNNs-based classification has shown impressive results for natural images, such as those in the ImageNet challenge dataset \cite{krizhevsky2012imagenet}, the handwritten digits (MNIST) dataset \cite{hadsell2006dimensionality} and the CIFAR-10 dataset \cite{he2016deep}. CNNs are also becoming popular in scientific and medical research, in areas such as tomography, magnetic resonance imaging, genomics, protein structure prediction etc. \cite{hua2014computer,wang2016perspective,park2015deep,zhou2014deep}. It is because of the proven success of CNNs that we chose to use one for EDS classification. 

In this paper, we first introduce the theory for SLADS and for detection and classification of EDS spectra. Then, we show results from four SLADS experiments performed on EDS data. In particular, we show experiments on a 2-phase sample measured at two different resolutions and experiments on a 4-phase sample measured at two different resolutions. We also evaluate the performance of our classifier.

\section{Theoretical Methods}

In this section we introduce the theory behind dynamic sampling as well as how we adapt it for EDS.

\subsection{SLADS Dynamic Sampling}
\label{SLADS}
Supervised learning approach for Dynamic Sampling (SLADS) was developed by Godaliyadda et al. \cite{Godaliyadda2,Godaliyadda3,Garth}.
The goal of dynamic sampling, in general, is to find the measurement which, when added to the existing dataset, has the greatest effect on the expected reduction in distortion (ERD). It is important to note that in this section we assume, as in the SLADS framework, that every measurement is a scalar quantity. We later elaborate how we generalize SLADS for EDS, where measurements are vectors. 

First, we define the image of the underlying object we wish to measure as $X \in \mathbb{R}^N$, and the value of location $s$ as $X_s$. Now assume we have already measured $k$ pixels from this image. Then we can construct a measurement vector,
$$
Y^{(k)} = 
\left[ 
\begin{array}{c}
s^{(1)}, X_{s^{(1)}} \\
\vdots \\
s^{(k)}, X_{s^{(k)}} 
\end{array}
\right] \ .
$$
Using $Y^{(k)}$ we can then reconstruct an image $\hat{X}^{(k)}$.

Second, we define the distortion between the ground-truth $X$ and the reconstruction $\hat{X}^{(k)}$ as $D \left(X,  \hat{X}^{(k)}\right)$. Here $D \left(X,  \hat{X}^{(k)}\right)$ can be any metric that accurately quantifies the difference between $X$ and $\hat{X}^{(k)}$. For example, if we have a labeled image, where each label corresponds to a different phase, then,
\begin{equation}
D \left(X,  \hat{X}^{(k)}\right) = \displaystyle\sum\limits_{i=1}^N I \left( X_i,\hat{X}^{(k)}_i \right),
\end{equation}
where $I$ is an indicator function defined as 
\begin{equation}
I \left( X_i,\hat{X}^{(k)}_i  \right) =\begin{cases}
0 \quad  X_i=\hat{X}^{(k)}_i \\
1 \quad X_i \neq \hat{X}^{(k)}_i.
\end{cases}
\end{equation}

Assume we measure pixel location $s$, where $s \in \left\lbrace \Omega \setminus \mathcal{S} \right\rbrace$, where $\Omega$ is the set containing indices of all pixels, and $\mathcal{S}$ is the set containing pixel locations of all measured pixels. Then we can define the reduction in distortion (RD) that results from measuring $s$ as,
\begin{eqnarray}
R^{(k;s)} = D ( X , \hat{X}^{(k)} ) - D ( X , \hat{X}^{(k;s)} ) \ .
\label{eqn:RD}
\end{eqnarray}
Ideally we would like to take the next measurement at the pixel that maximizes the RD. However, because we do not know $X$, i.e. the ground-truth, the pixel that maximizes the expected reduction in distortion (ERD) is measured in the SLADS framework instead. The ERD is defined as, 
\begin{equation}
\bar{R}^{(k;s)}= \mathbb{E} \left[ R^{(k;s)} \vert Y^{(k)} \right] \ .
\label{eqn:ERD}
\end{equation}
Hence, in SLADS the goal is to measure the location, 
\begin{equation}
s^{(k+1)}= \arg \max_{s \in \Omega } \left\lbrace \bar{R}^{(k;s)} \right\rbrace.
\label{eqn:ideal sampling strategy}
\end{equation}

In SLADS the relationship between the measurements and the ERD for any unmeasured location $s$ is assumed to be given by, 
\begin{equation}
\mathbb{E} \left[ R^{(k;s)} \vert Y^{(k)} \right] \ = \hat{\theta} V^{(k)}_s.
\end{equation}
Here, $V^{(k)}_s$ is a $t \times 1$ feature vector extracted for location $s$ and $\hat{\theta}$ is $1 \times t$ vector that is computed in training. The training procedure is detailed in \cite{Godaliyadda2,Godaliyadda3} and therefore will not be detailed here. 

\subsection{Adapting SLADS for Energy-Dispersive Spectroscopy}
In SLADS it is assumed that a measurement is a scalar value. However, in EDS, the measurement spectrum is a $p \times 1$ vector. So to use SLADS for EDS we either need to redefine the distortion metric, or convert the $p \times 1$ vector spectra to a labeled discrete class of scalar values. In this paper, we use the latter approach.

\begin{figure}
\centering
\subcapraggedrighttrue
\subfigure[{ \scriptsize }]{\includegraphics[scale=0.5]{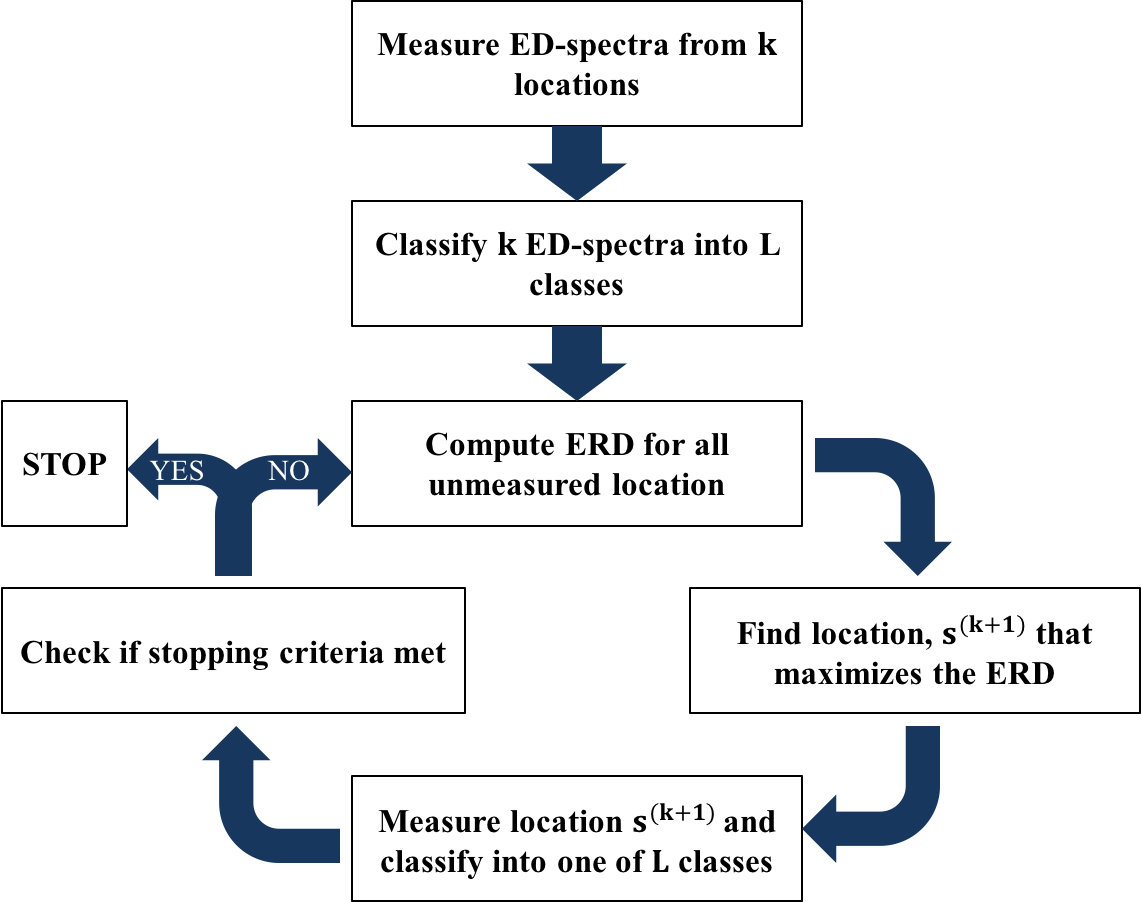}} 
\subfigure[{ \scriptsize }]{\includegraphics[scale=0.5]{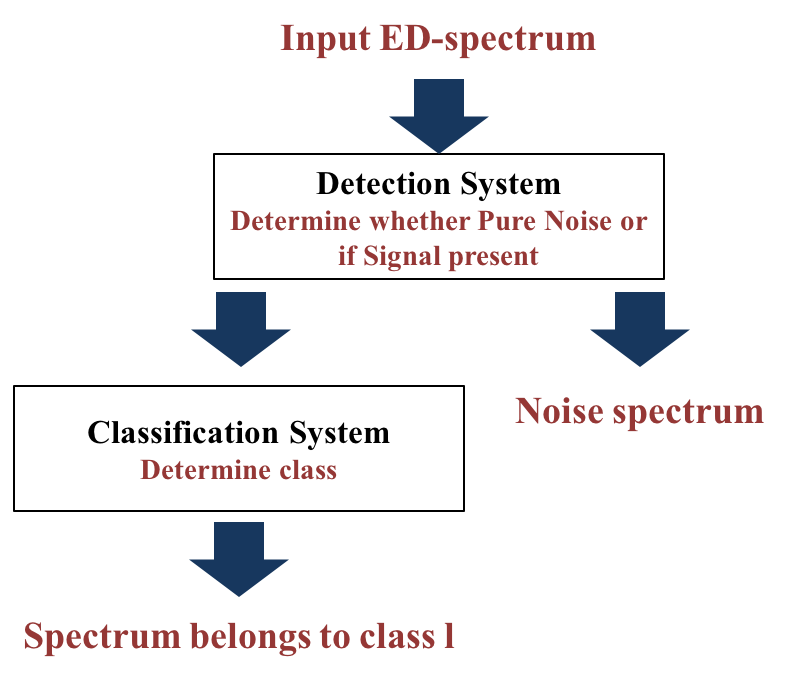}}\\
\caption{(a) The SLADS algorithm adapted for ED-spectra and (b) The two-tiered structure of our EDS classification system.}
\label{fig:SLADS for EDS}
\end{figure}

In order to make a meaningful conversion, the scalar value should be descriptive of the measured energy spectrum, and ultimately allow us to obtain a complete understanding of the underlying object. The objective in this work is to identify the distribution of different phases in the underlying object. Note that a phase is defined as the set of all locations in the image that have the same EDS spectrum. So if we can classify the measured spectra into one of $L$ classes, where the $L$ classes correspond to the $L$ different phases, and hence $L$ different spectra, then we can readily adapt SLADS for EDS. Figure \ref{fig:SLADS for EDS}(a) shows this adaptation in more details. The method we used for classifying spectra is detailed in the next section.

\subsection{Classifying Energy Dispersive Spectra}
\label{Classifying EDS}

Assume the EDS measurement from a location $s$ is given by $Z_s \in \mathbb{R}^p$, where $p>1$. Hence we need to convert $Z_s$ to a discrete integer class $X_s$, where $X_s$ is a label that corresponds to the elemental composition (phase) at location $s$. Also, let us assume that the sample we are measuring has $L$ different phases, and therefore, $X_s \in \left\lbrace 0,1, \hdots L \right\rbrace$, where $0$ corresponds to an \textit{ill-spectrum}. An \textit{ill-spectrum} can be caused by a sample defect, equipment noise or other undesired phenomena, and therefore is not from one of the $L$ phases that are known. 

In this paper, to classify $Z_s$ in to one of $L+1$ classes we use a two step approach as shown in Figure \ref{fig:SLADS for EDS}(b). In the first step, we determine if the measured spectrum is an ill-spectrum. We call this step the \textit{detection step}. If we determine that $Z_s$ is an ill-spectrum then we let $X_s=0$. If not we move on to the second step of determining which of the $L$ phases $Z_s$ belongs to and assign that label to $X_s$. We call this second step the \textit{classification step}. In the next two sections we will explain the algorithms we used for the Detection and Classification steps.

\subsubsection{Detection using Neural Network Regression}

In the detection step, we use a neural network regression (NNR) model to detect the ill-spectrum class \cite{specht1991general}. The NNR model we use has $Q$ neurons in each hidden layers as well as the output layer. 

Assume that we have $M$ training spectra for each of the $L$ phases. The goal of training is to find a function $\hat{f}(\cdot)$ that minimizes the Loss function and project training spectra onto a pre-set straight line $f$, where:
\begin{equation}
Loss = \frac{1}{2} \sum_{r \in \left\lbrace 1,2,\hdots, LM \right\rbrace} || f -  \hat{f}(Z_s) ||^2
\end{equation}
Here, $LM$ is the total number of training spectra and $Z_s$ where $s \in \left\lbrace 1,2,\hdots, LM \right\rbrace$ denotes one training spectrum. It is important to note that since this is a neural network architecture, by saying we find $\hat{f}(\cdot)$, it is understood that we find the weights of the neural networks, that correspond to $\hat{f}(\cdot)$.

To determine if a spectrum $Z_s$ is an ill-spectrum or one which belongs to one of $L$ phases, we first compute,
\begin{equation}
g_s = \vert f - \hat{f}(Z_s) \vert.
\end{equation}
where, $g_s \in \mathbb{R}^Q$. Then we compute the variance metric, 
\begin{equation}
\sigma^2 \left( Z_s \right) = \frac{1}{Q} \sum_{i=1}^{Q} \left[ g_{s,i} - \mu_s \right]^2
\end{equation}
where, $g_{s,i}$ is the $i^{th}$ element of the vector $g_s$ and 
\begin{equation}
\mu_s = \frac{1}{Q} \sum_{i=1}^Q g_{s,i}
\end{equation}
Then a pre-set threshold is applied to the variance metric to decide whether the $Z_s$ is a ill-spectrum i.e.
\begin{eqnarray}
\hat{X}_s =
\begin{cases}
0,  \hfill   \sigma^2(Z_s) > T \\
\left\lbrace 1,2,\hdots L \right\rbrace, \hfill  \sigma^2(Z_s) \leq T.
\end{cases}
\end{eqnarray}

\subsubsection{Classification using Convolutional Neural Networks}
The next task at hand is to classify the spectrum according to one of the $L$ labels, given that we found a spectrum $Z_s$ for some location $s$, which is not an ill-spectrum. For this classification problem we use the convolutional neural networks (CNNs) as described in \cite{krizhevsky2012imagenet} and implement using tensorflow \cite{abadi2016tensorflow}.

\begin{figure}
\centering
\includegraphics[scale=0.2]{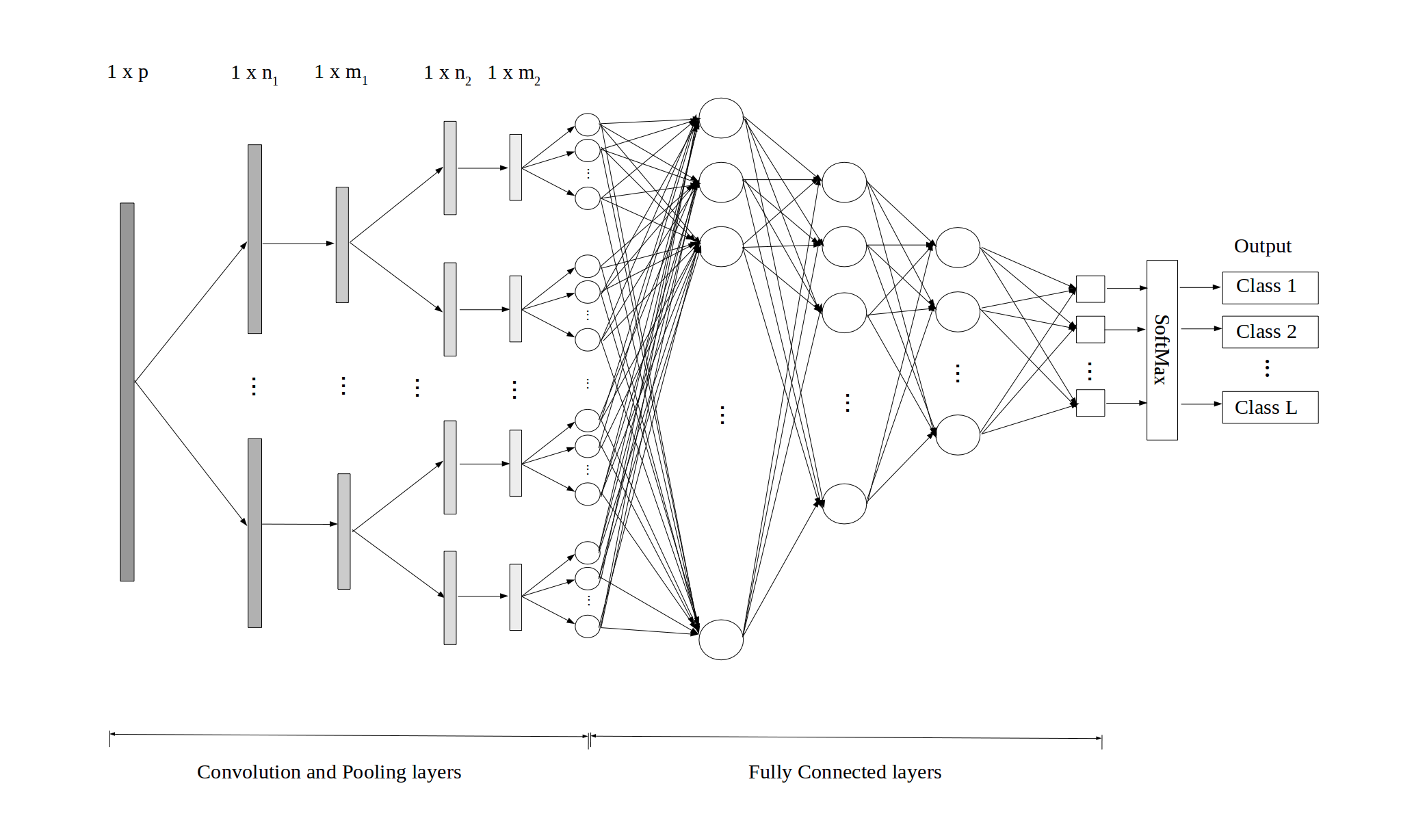}
\caption{The CNNs architecture for EDS spectra classification.}
\label{fig:CNNs}
\end{figure}

The CNNs we use in this paper has two convolution layers, each followed by a max-pooling layer followed by three fully connected layers, shown in Figure \ref{fig:CNNs}.
The first convolution layer has $u_1$ $1 \times k$ kernels sliding across the input spectrum with a stride $v$ to extract $u_1$ features, each of size $1 \times n_1$. 
The max-pooling layer that follows this layer again operates with the same stride and kernel size, resulting in $u_1$ features, each of size $1 \times m_1$. 
The second convolution layer increases the number of features from $u_1$ to $u_2$, where, $u_2 \mod u_1 =0$, by the application of $u_2$ kernels of size $1 \times k$ at the same stride ($v$) to the output of the first max-pooling layer. 
The max pooling layer that follows is identical to the previous max-pooling layer.

After the convolution and max-pooling layers, all feature values are stacked into a single vector known as a \textit{flat layer}. The flat layer is the transition into the fully connected layers that follow. 
These layers have the same architecture as typical neural networks. 
In the fully connected layers, the number of neurons in each layer is reduced in our implementation.  

The output of the fully connected layers is a $1 \times L$ vector, $X_s^{1}$. Each entry of this vector is then sent through a SoftMax function to create again an $1 \times L$ vector, which we will denote as $X_s^{2}$, for a location $s$.
\begin{equation}
X_{s,i}^{2} = \frac{\exp \left( X_{s,i}^{1} \right)}{\sum_{j=1}^{L} \exp \left( X_{s,j}^{1} \right) }.
\end{equation}
Here, $X_{s,j}^{1}$ corresponds to the $j^{th}$ component of $X_{s}^{1}$.

Now assume we have the same training examples as in the previous section i.e. $M$ spectra from the $L$ phases. When training the CNNs we minimize the Cross-Entropy, defined as,
\begin{equation}
CE = - \sum_{r=1}^{LM} X_s^{one-hot} \log X_s^{2}  
\end{equation}
where, $LM$ is the total number of training samples, and $X_s^{one-hot}$ is the ``one-hot" representation of $X_s$. Here, the ``one-hot" notation of label $m$, when $L$ labels are available, is a $L \times 1$ dimensional vector with $1$ at location $m$ and zeros everywhere else.

\section{Experimental Methods and Results}
In this section, we will first describe the experimental methods used in the simulated experiments in Section \ref{Experimental Methods} and then present the results for the simulated experiments in Section \ref{Results}.

\subsection{Experimental Methods}
\label{Experimental Methods}
Here we first present how we generate images for training and \textit{simulated objects} to perform SLADS on and finally how we train the classifier we described in Section \ref{Classifying EDS}. For all the experiments we used a Phenom ProX Desktop SEM. The acceleration voltage of the microscope was set to 15 kV. 
To acquire spectra for the experiments, we used the EDS detector in \textit{spot mode} with an acquisition time of 10 s.  

\subsubsection{Constructing Segmented Images for Training}
We first acquired representative images from the object with $L$ different phases using the back scattered detector on the Phenom. Then we segmented this image so that each label would correspond to a different phase. Finally we denoised the image using an appropriate denoising scheme to create a clean image. 

\subsubsection{Constructing a Simulated Object}
\label{create object}
In this paper, we want to dynamically measure an object in the EDS mode. This means that if the beam is moved to a location $s$, the measurement extracted, $Z_s$, is a $p$ dimensional vector. So we can think of the problem as sampling an object with dimensions $N \times N \times p$, where we can only sample sparsely in the spatial dimension, i.e. the dimension with $N \times N$ points. This hypothetical object is what we call here a\textit{ simulated object}.

To create the\textit{ simulated object}, we first acquire an SEM image, and segment it just as in the previous section. We then add noise to this image by assigning the label $0$ to a randomly selected set of the pixels. Then we experimentally collect $M$ different spectra from each of the $L$ different phases using the EDS detector in \textit{spot mode}. Finally we raster through the segmented and noise added image and assign a spectrum to each pixel location in the following manner: If the value read at a pixel location $s$ is $0$ we assign a pure noise spectrum to that location. If the value read is $l \in \left\lbrace 1, 2, \hdots L \right\rbrace$, we randomly pick one of the $M$ spectra we acquired previously for phase $l$ and add Poisson noise to it. Then we assign the noise added spectrum to location $s$. It is important to note that the noise added to each location is independent of location and is different for each pixel location. So now we have an object of size $N \times N \times p$ to use in our SLADS experiment.  

\subsubsection{Specification of the Neural Networks to Classify EDS Spectra}
\label{Train NN}
In order to train and validate the detection and classification neural networks, we again collected $M^{train}$ spectra for each phase. Then we added Poisson noise to each spectrum and then used half of the spectra to train the NNR and the CNNs, and used the other half to validate, before using it in SLADS.

The NNR network we used has $5$ fully connected hidden layers, each with $100$ neurons. The CNNs classification system we used has kernel size $k = 10$ and a stride of $2$ for all convolution and max-pooling layers. The number of features in the first and second convolution layers are $8$ and $16$ respectively. The flat layer stacks all features from second max-pooling layer into a $2048$ dimensional vector. The number of neurons for the following $3$ fully connected layers are $100$, $32$ and $8$. 

\subsection{Results}
\label{Results}
In this section, we will present results from SLADS experiments performed on $4$ different simulated objects. To quantify the performance of SLADS, we will use the total distortion (TD) metric.  
The TD after $k$ measurements are made is defined as, 
\begin{equation}
TD_k = \frac{1}{\vert \Omega \vert} D \left( X,\hat{X}^{(k)} \right).
\end{equation}
We will also evaluate the accuracy of the classification by computing the misclassification rate. 

\subsubsection{Experiment on Simulated 2-Phase Object with Pb-Sn Alloy}
Here we will present results from sampling two $2$-phase simulated objects, one with dimensions $128 \times 128 \times p$, and the other with dimensions $1024 \times 1024 \times p$ created using spectra and SEM images acquired from a Pb-Sn eutectic alloy sample \cite{rowenhorst2006three}.
It is important to note that the dimension $p$ here corresponds to the dimension of the spectrum, i.e. in the simulated object at each pixel location we have a $p$ dimensional spectrum.
One of the phases has Pb and Sn, and the other only Sn. 

\begin{figure}
\includegraphics[scale=0.75]{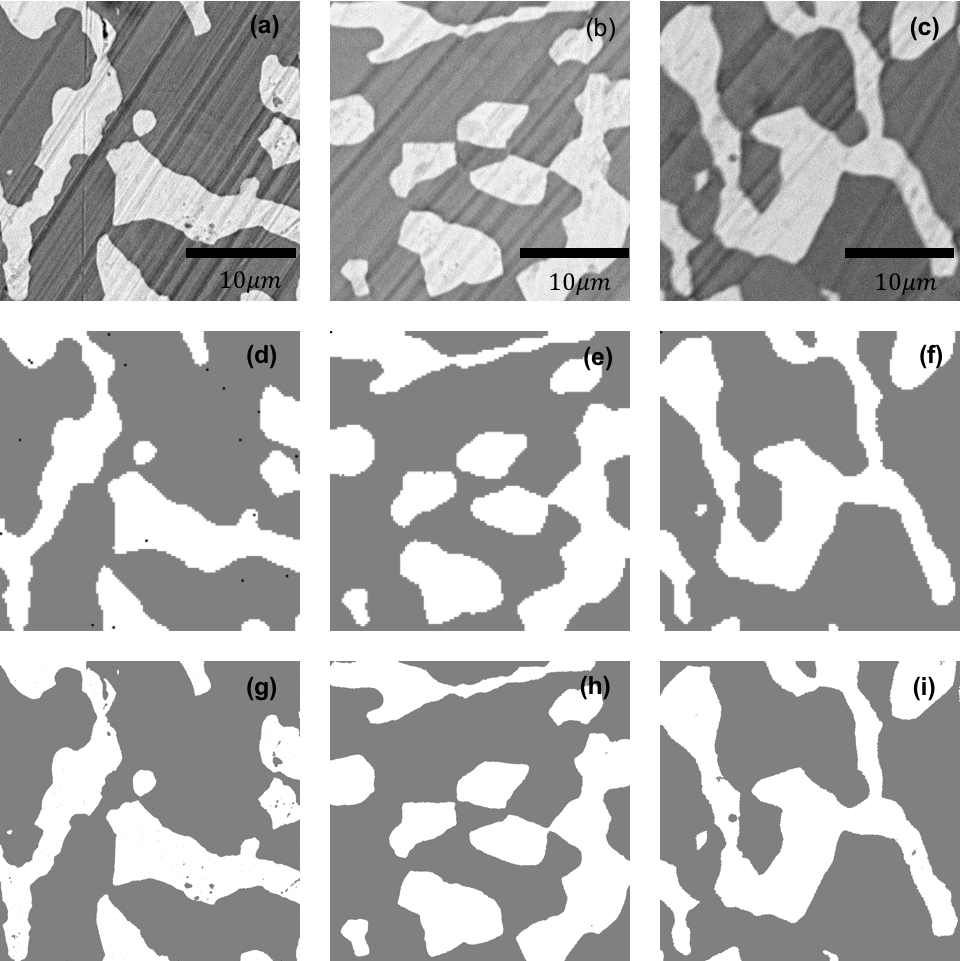}
\caption{Testing and training Images for the $2$-Phase SLADS experiments. (a)-(c): original SEM images; (d)-(f): $128 \times 128$ images with labels $0,1,$ and $2$; (g)-(i): $1024 \times 1024$ images with labels $0,1,$ and $2$; The images in the first column are the ones used for testing and the others are the ones used for training.}
\label{fig:Experiment_12_Images}
\end{figure}

Both these objects, as well as the training data for SLADS, were created using SEM images taken at  $1024 \times 1024$ resolution. We created $128 \times 128$ images by down-sampling the original $1024 \times 1024$ images. Then we used a simple thresholding scheme to segment all the SEM images. Then we added noise only to the testing images. The images we used for testing and training are shown in Figure \ref{fig:Experiment_12_Images}.

To train and validate the neural networks we acquired $24$ spectra from each phase. To create the simulated testing object we acquired and used $12$ (different) spectra for each phase. The noise added to the spectra while creating the simulated object is Poisson noise with $\lambda = 2$. The ill-spectrum we generated were also Poisson random vectors, with independent elements and $\lambda = 20$.

\begin{figure}
\centering
\subcapraggedrighttrue
\subfigure[{ \scriptsize Measurement Locations $15 \%$}]{\includegraphics[height=1.1in,width=1.1in]{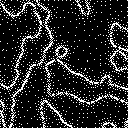}} 
\subfigure[{ \scriptsize Reconstructed Image $15 \%$}]{\includegraphics[height=1.1in,width=1.1in
]{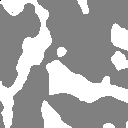}}
 \subfigure[{ \scriptsize Ground-Truth Image\hspace{5mm} }]{\includegraphics[height=1.1in,width=1.1in
]{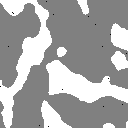}}
 \subfigure[{ \scriptsize Distortion Image $15 \%$}]{\includegraphics[height=1.1in,width=1.1in
]{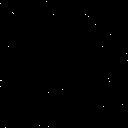}}\\

\caption{Results of EDS-SLADS experiment performed on the $2$-phase simulated object of size $128 \times 128 \times 2040$. Here we have acquired $15 \%$ of available measurements.}
\label{fig:Experiment_1}
\end{figure}

The results after $15\%$ of samples were collected from the $128 \times 128$ image is shown in Figure \ref{fig:Experiment_1}. The TD with $15\%$ of samples was $0.0015$. From this figure it is clear that we can achieve near perfect reconstruction with just $15\%$ of samples. The misclassification rate of the detection and classification system was computed to be $0.0002$, which tells us that the detection and classification system is also very accurate.

\begin{figure}
\centering
\subcapraggedrighttrue
\subfigure[{ \scriptsize Measurement Locations $5\%$}]{\includegraphics[height=1.1in,width=1.1in]{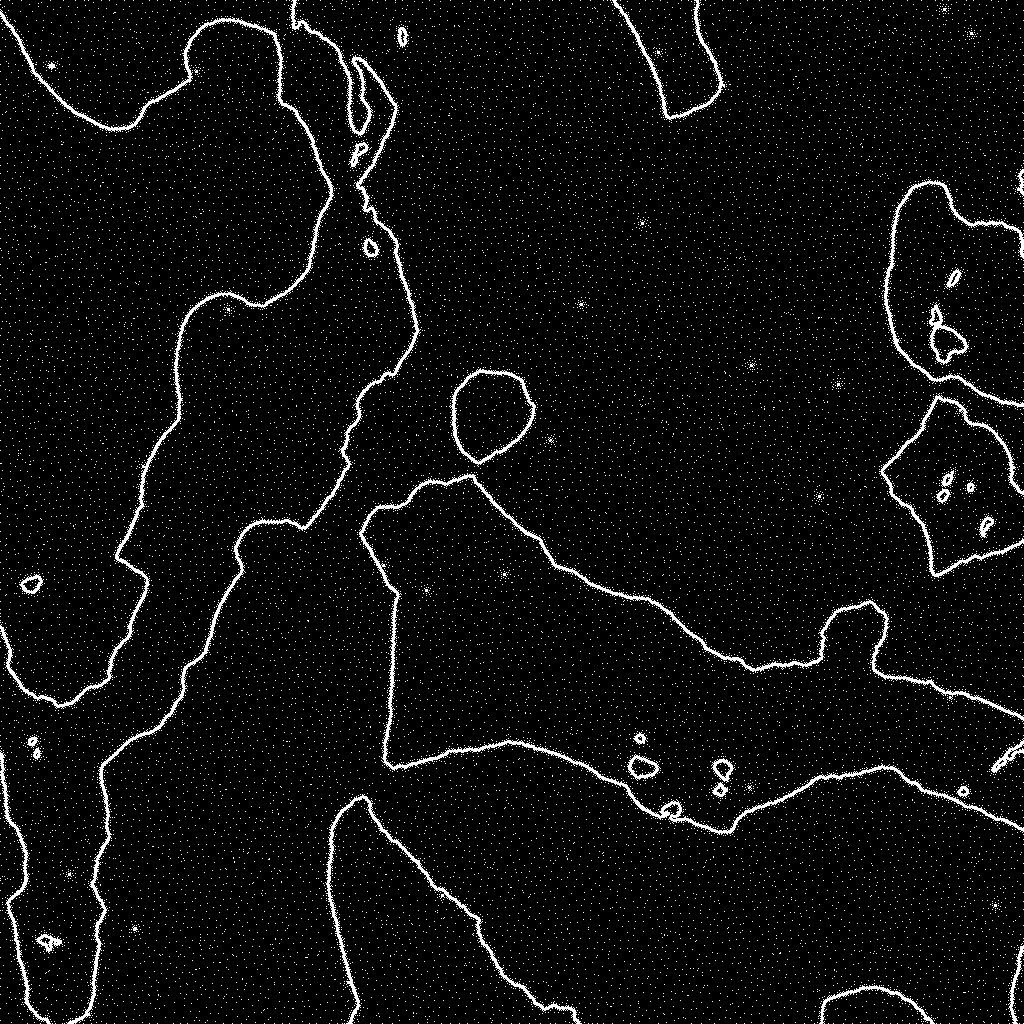}} 
\subfigure[{ \scriptsize Reconstructed Image $5\%$}]{\includegraphics[height=1.1in,width=1.1in
]{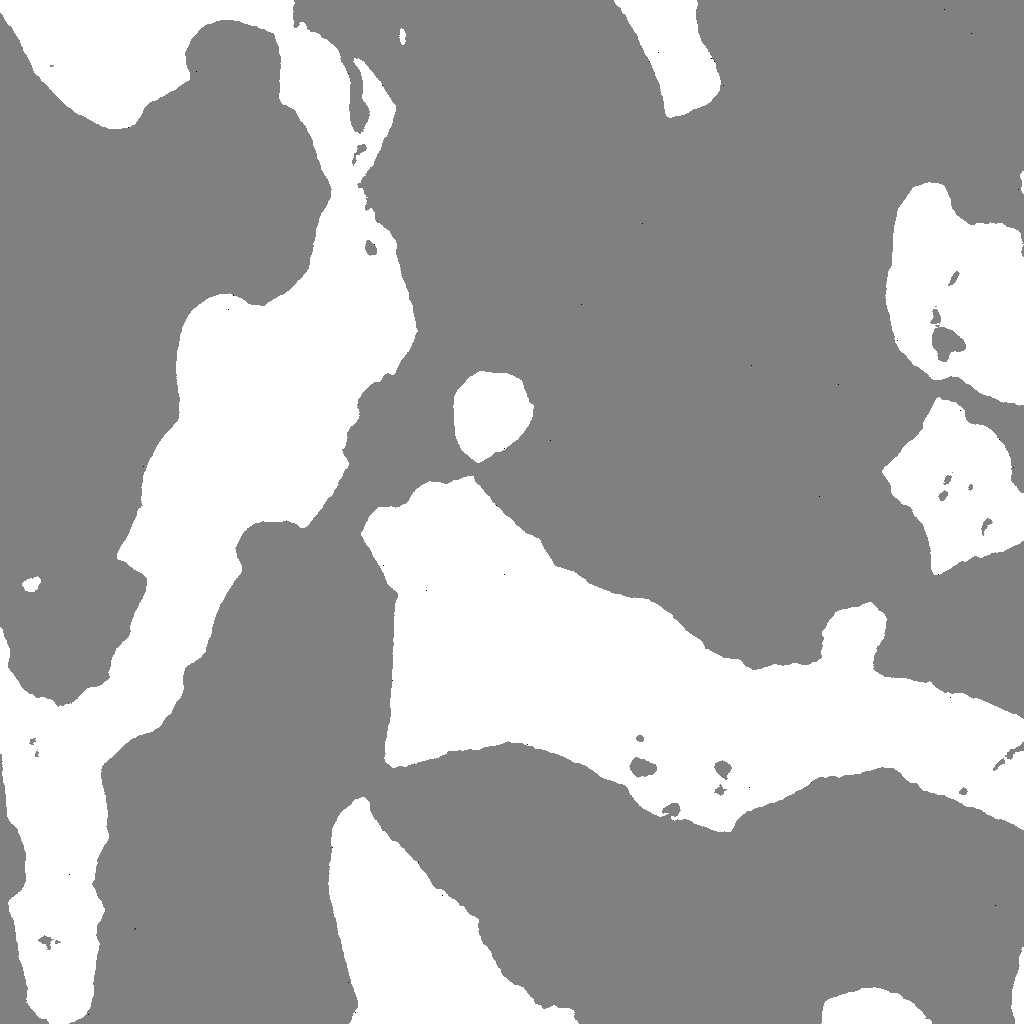}}
 \subfigure[{ \scriptsize Ground-Truth Image\hspace{5mm} }]{\includegraphics[height=1.1in,width=1.1in
]{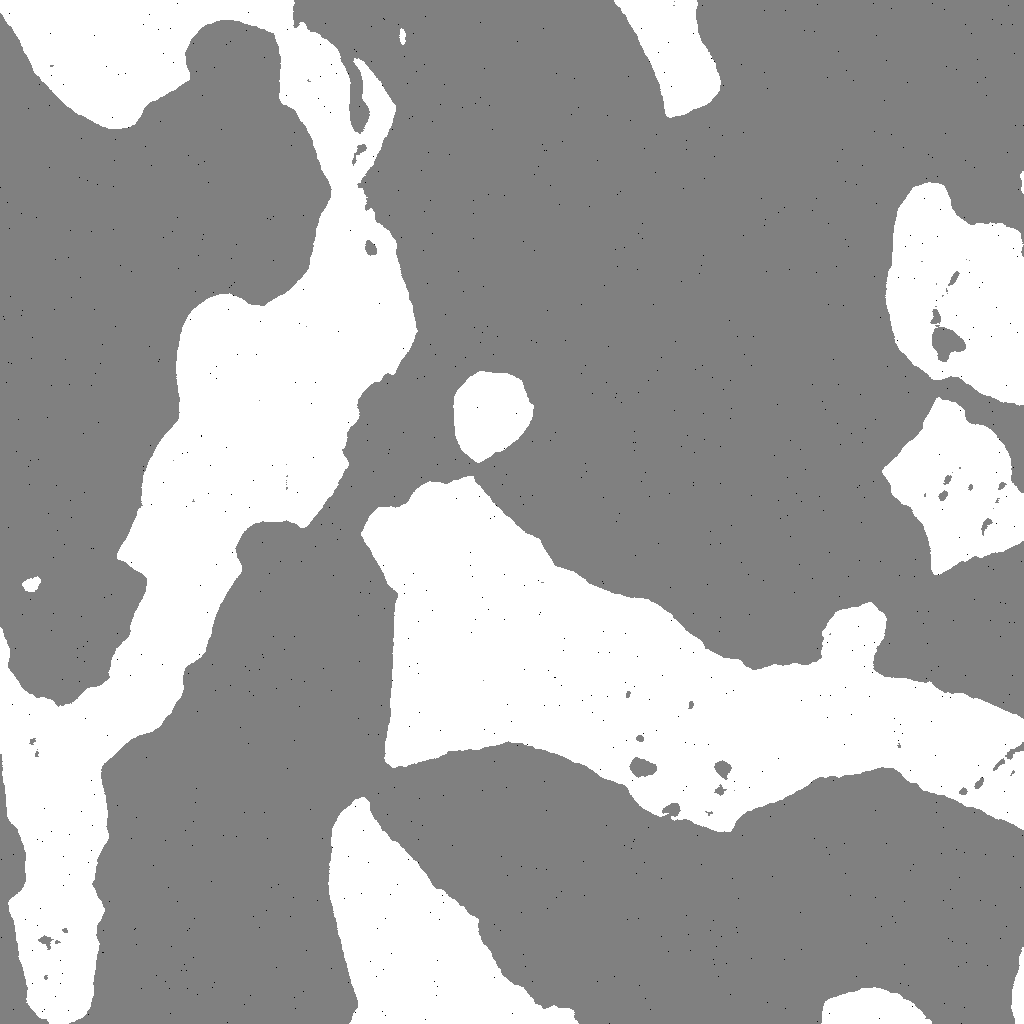}}
 \subfigure[{ \scriptsize Distortion Image $5\%$ }]{\includegraphics[height=1.1in,width=1.1in
]{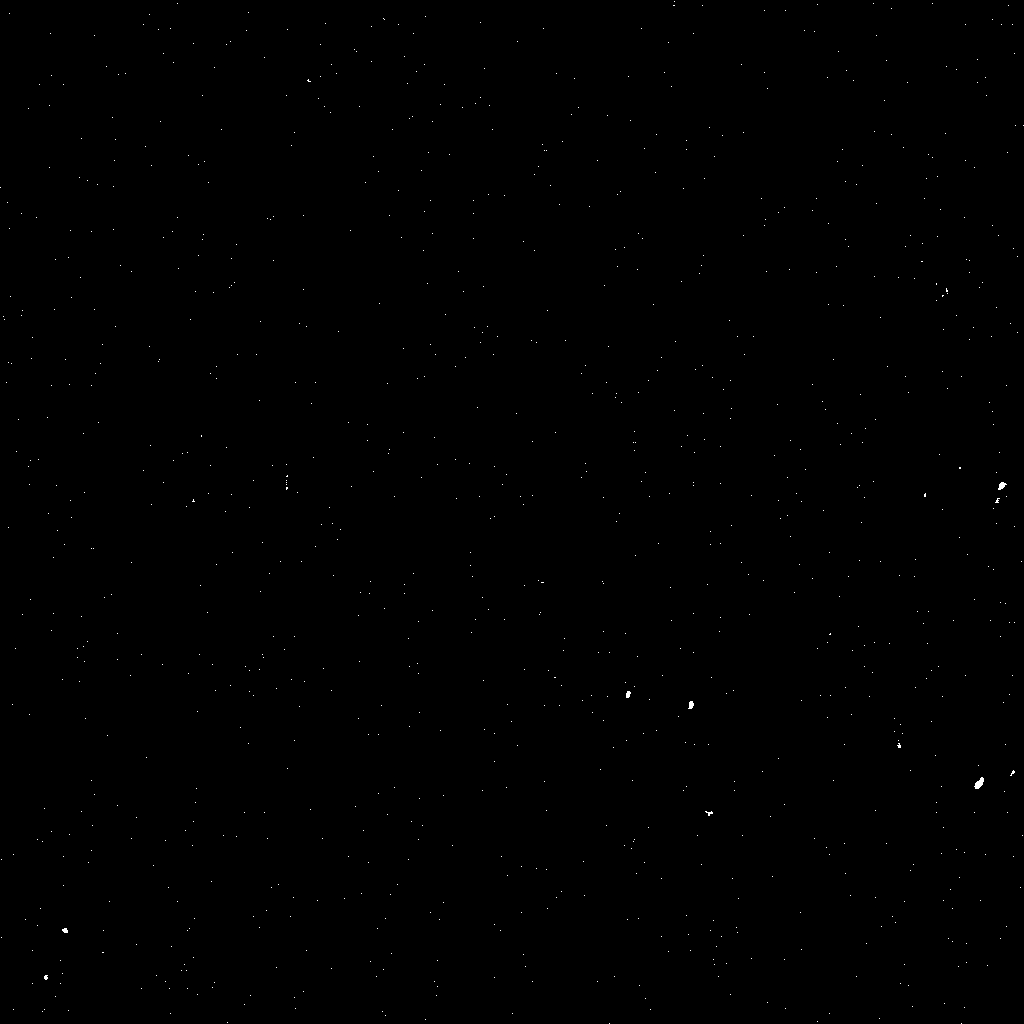}}\\

\caption{Results of EDS-SLADS experiment performed on the $2$-phase simulated object of size $1024 \times 1024 \times 2040$. Here we have acquired $5 \%$ of available measurements.}
\label{fig:Experiment_2}
\end{figure}

The results after $5\%$ of samples were collected from the $1024 \times 1024$ image is shown in Figure \ref{fig:Experiment_2}. The TD with $5\%$ of samples was $0.0013$.
Here we see that even for the same object, if we sample at a higher spatial resolution we can achieve similar results with just $5\%$ of measurements. In this experiment the misclassification rate of the detection and classification system was computed to be $0$. 

\subsubsection{Experiment on Simulated 4-phase Object}
For this experiment we will again sample two simulated objects, one with dimensions $256 \times 256 \times p$, and the other with dimensions $1024 \times 1024 \times p$ created using spectra and SEM images from a micro-powder mixture with $4$ phases i.e. CaO, LaO,Si and C.

\begin{figure}
\includegraphics[scale=0.75]{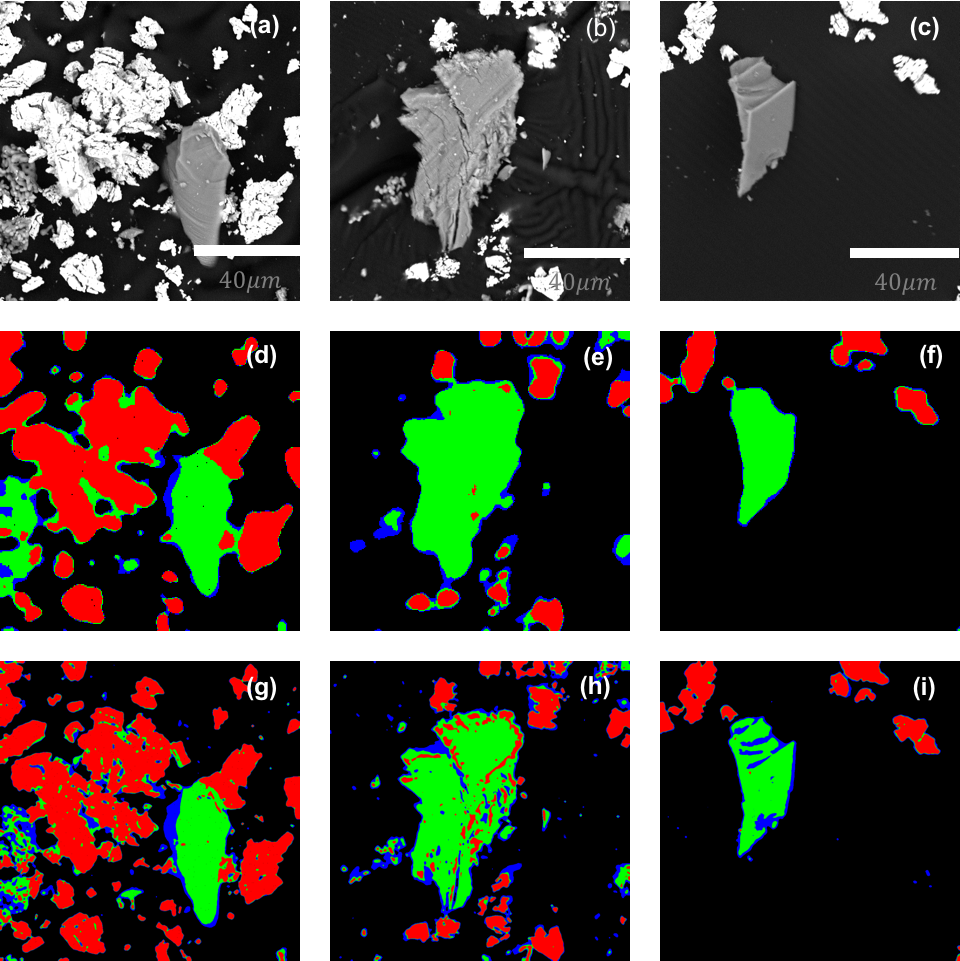}
\caption{Testing and training Images for the $4$-Phase SLADS experiments. (a)-(c): original SEM images; (d)-(f): $256 \times 256$ images with labels $0,1,2,3,$ and $4$; (g)-(i): $1024 \times 1024$ images with labels $0,1,2,3,$ and $4$; The images in the first column are the ones used for testing and the others are the ones used for training.}
\label{fig:Experiment_34_Images}
\end{figure}

The testing and training images, shown in Figure \ref{fig:Experiment_34_Images} were created in exactly same manner as in the previous experiment. The testing object was again created with $12$ spectra from each phase. The neural networks were also trained validated just as before once more using $24$ spectra from each phase.

\begin{figure}
\centering
\subcapraggedrighttrue
\subfigure[{ \scriptsize Measurement Locations $20\%$}]{\includegraphics[height=1.1in,width=1.1in]{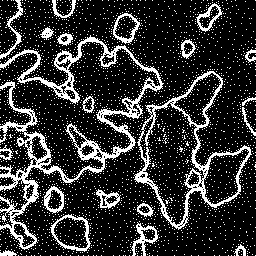}} 
\subfigure[{ \scriptsize Reconstructed Image $20\%$}]{\includegraphics[height=1.1in,width=1.1in
]{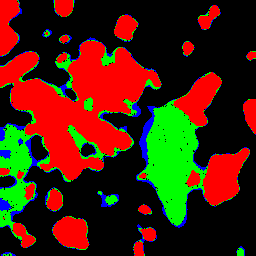}}
 \subfigure[{ \scriptsize Ground-Truth Image\hspace{5mm} }]{\includegraphics[height=1.1in,width=1.1in
]{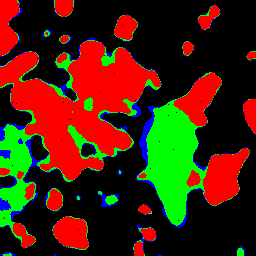}}
 \subfigure[{ \scriptsize Distortion Image $20\%$ }]{\includegraphics[height=1.1in,width=1.1in
]{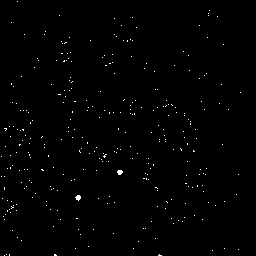}}\\

\caption{Results of EDS-SLADS experiment performed on the $4$-phase simulated object of size $256 \times 256 \times 2040$. Here we have acquired $20 \%$ of available measurements.}
\label{fig:Experiment_3}
\end{figure}

The results after $20\%$ of samples were collected from the $256 \times 256$ image is shown in Figure \ref{fig:Experiment_3}. The TD with $20\%$ of samples was $0.006$. Again we see that we can achieve near perfect reconstruction with $20\%$ of samples. The misclassification rate of the detection and classification system was computed to be $0.005$, which again tells us that the detection and classification system is very accurate. However, we do note that this is not as accurate as in the $2$-phase case.

\begin{figure}
\centering
\subcapraggedrighttrue
\subfigure[{ \scriptsize Measurement Locations $5\%$}]{\includegraphics[height=1.1in,width=1.1in]{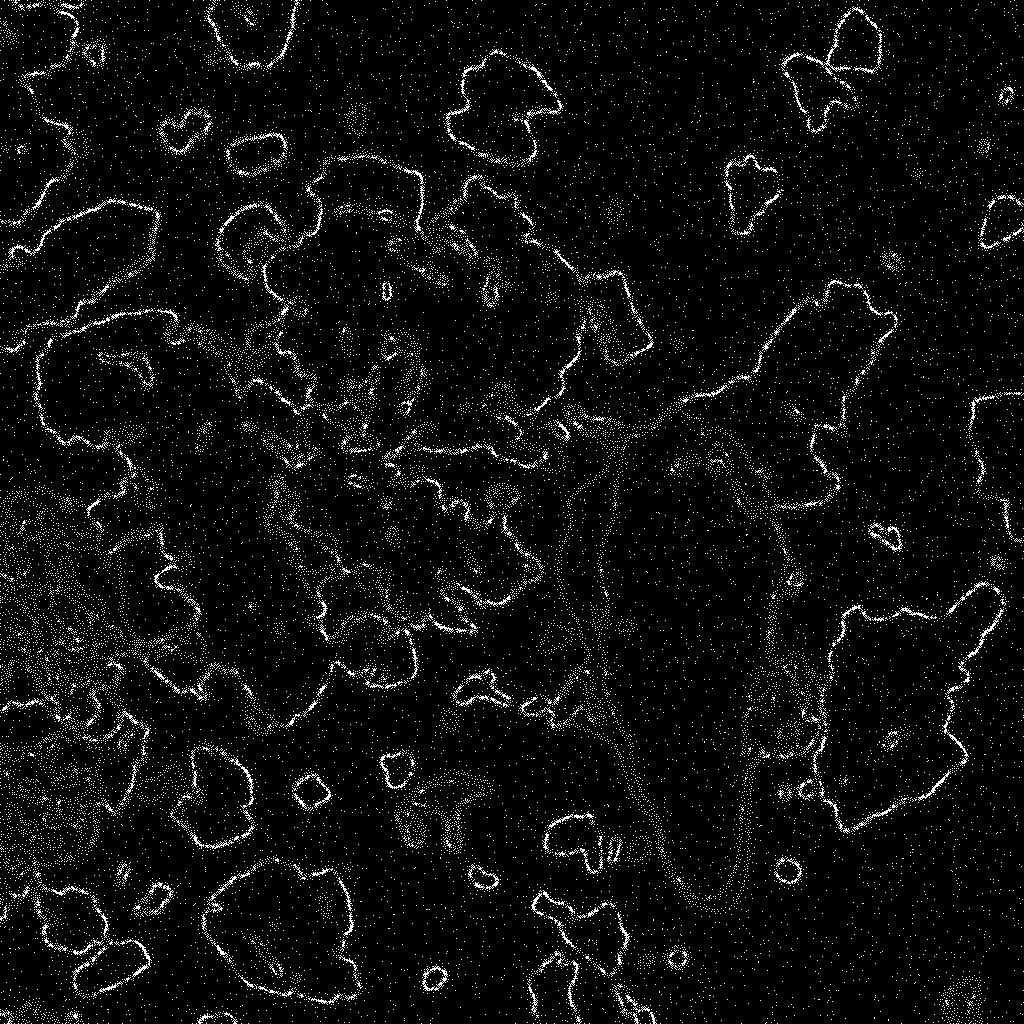}} 
\subfigure[{ \scriptsize Reconstructed Image $5\%$}]{\includegraphics[height=1.1in,width=1.1in
]{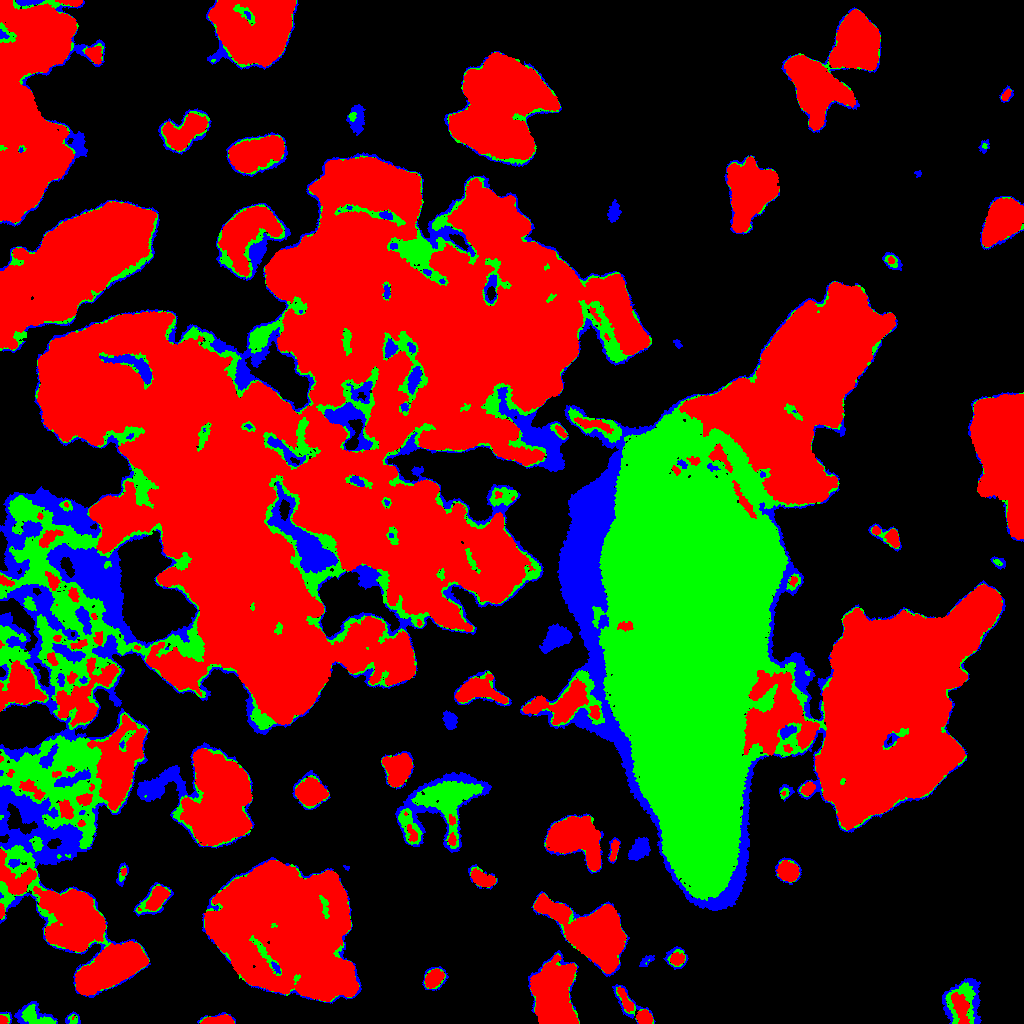}}
 \subfigure[{ \scriptsize Ground-Truth Image\hspace{5mm} }]{\includegraphics[height=1.1in,width=1.1in
]{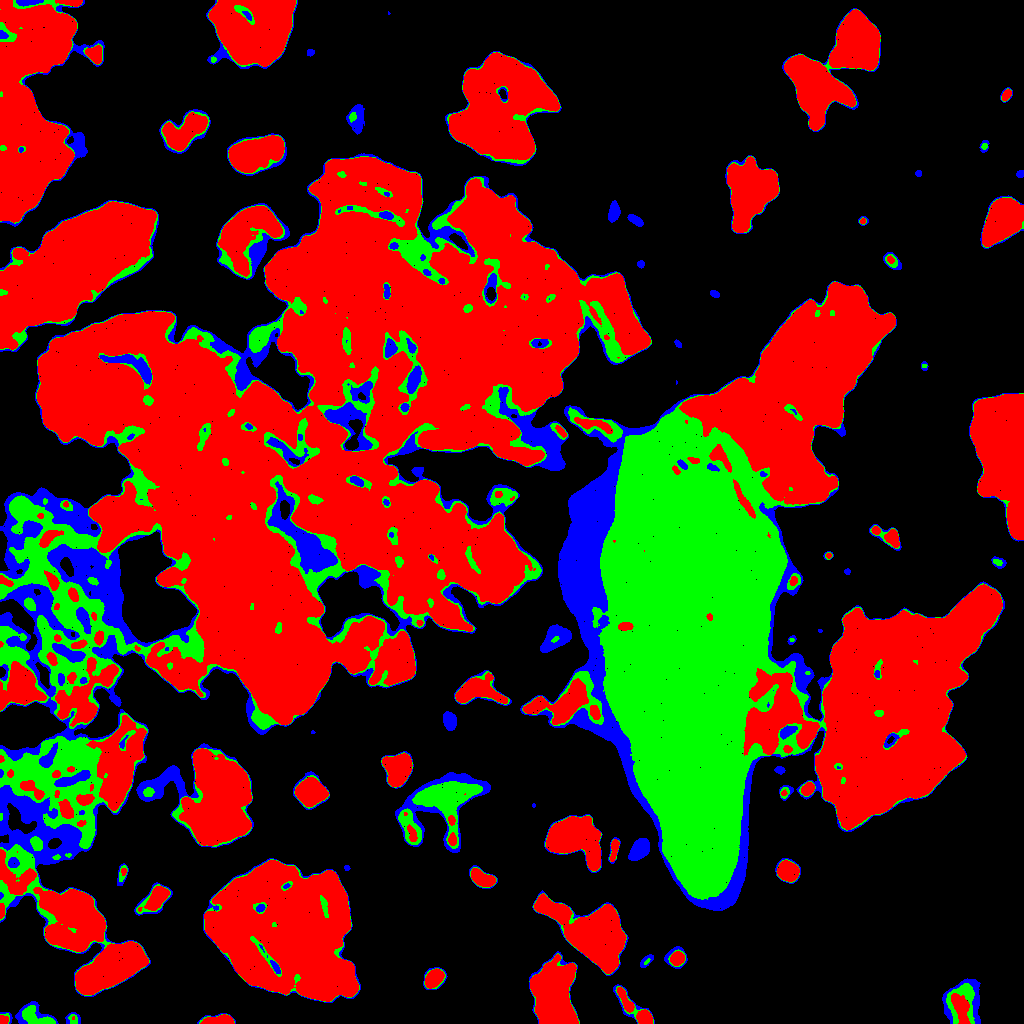}}
 \subfigure[{ \scriptsize Distortion Image $5\%$ }]{\includegraphics[height=1.1in,width=1.1in
]{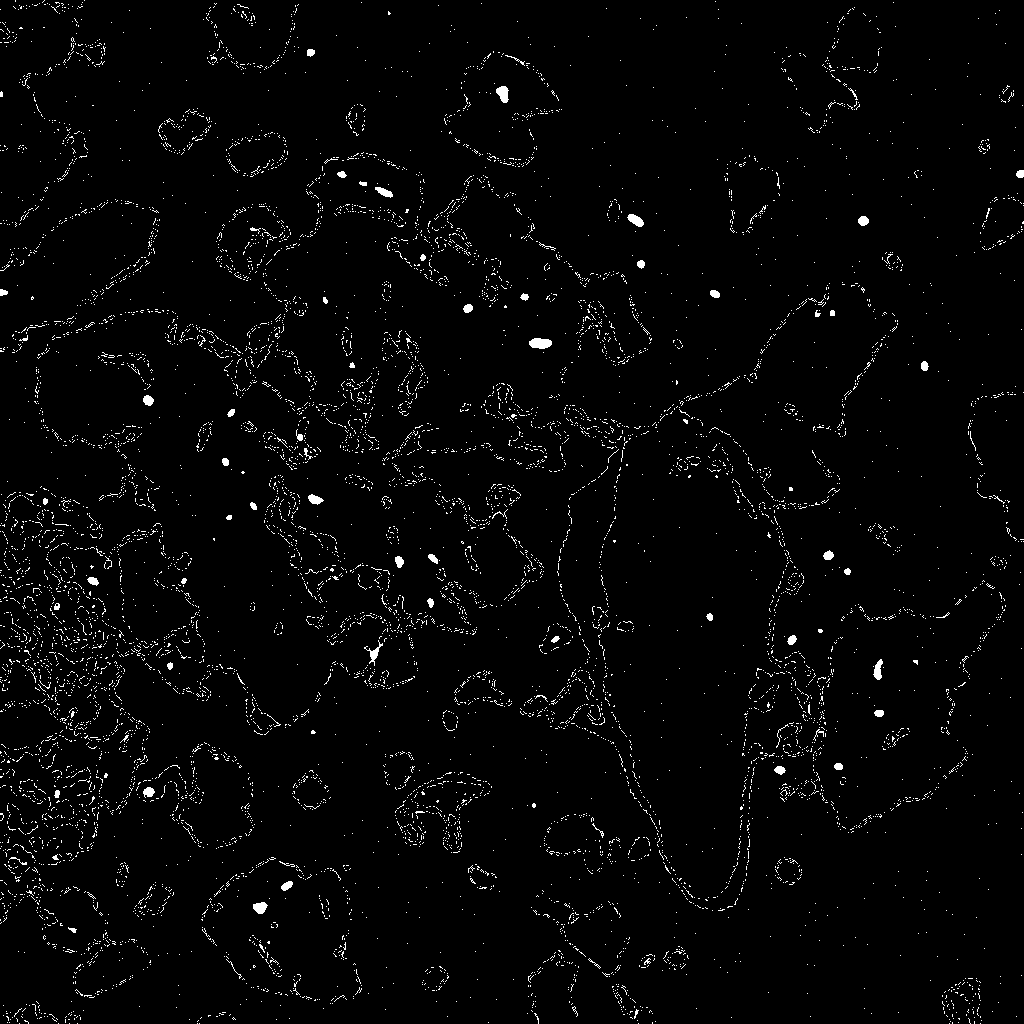}}\\

\caption{Results of EDS-SLADS experiment performed on the $4$-phase simulated object of size $1024 \times 1024 \times 2040$. Here we have acquired $5 \%$ of available measurements.}
\label{fig:Experiment_4}
\end{figure}

The results after $5\%$ of samples were collected from the $1024 \times 1024$ image is shown in Figure \ref{fig:Experiment_4}. The TD with $5\%$ of samples was $0.02$. In this experiment the misclassification rate of the detection and classification system was computed to be $0.0009$. 

\section{Discussion}

It is clear that by using SLADS to determine the sampling locations, we can reduce the overall exposure to anywhere between $5-20\%$, and still obtain a near perfect reconstruction. These results also show that the SLADS method is better suited for higher pixel resolution mapping which maximizes the resolution capability of the instrument and detector. 
This method would be useful for investigation of biological and beam-sensitive samples, such as live cell imaging, as well as for high-throughput imaging of large samples, such as fabrication by additive manufacturing and defects metrology in chemical and structural study.

\section{Conclusion}
In conclusion, we have shown that integrating dynamic sampling (SLADS) with EDS classification (CNNs) offers significant advantage in terms of dose reduction and the overall data acquisition time. We have shown that when imaging at lower pixel resolution i.e. $128 \times 128$ or $256 \times 256$ we can achieve a high-fidelity reconstructions with approximately $20\%$ samples and when imaging at higher pixel resolution i.e $1024 \times 1024$ we can achieve a high-fidelity reconstruction with just $5\%$ samples. We have also shown that our classification algorithm performs remarkably well in all the experiments. 
For future work, we will expand our EDS training database by including analytically simulated EDS data for more commonly used elements. We will use pure simulated EDS data to train the classification system, which enables more automated, high-throughput acquisition and characterization across different microscopic and spectroscopic platforms.

\section{Acknowledgement}
This material is based upon work supported by Laboratory Directed Research and Development (LDRD) funding from Argonne National Laboratory, provided by the Director, Office of Science, of the U.S. Department of Energy under Contract No. DE-AC02-06CH11357.


\begin{thebibliography}{10}
\expandafter\ifx\csname url\endcsname\relax
  \def\url#1{\texttt{#1}}\fi
\expandafter\ifx\csname urlprefix\endcsname\relax\def\urlprefix{URL }\fi
\expandafter\ifx\csname href\endcsname\relax
  \def\href#1#2{#2} \def\path#1{#1}\fi

\bibitem{goldstein2012scanning}
J.~Goldstein, D.~E. Newbury, P.~Echlin, D.~C. Joy, A.~D. Romig~Jr, C.~E. Lyman,
  C.~Fiori, E.~Lifshin, Scanning electron microscopy and X-ray microanalysis: a
  text for biologists, materials scientists, and geologists, Springer Science
  \& Business Media, 2012.

\bibitem{Allen2010}
A.~J. D'Alfonso, B.~Freitag, D.~Klenov, L.~J. Allen,
  \href{https://link.aps.org/doi/10.1103/PhysRevB.81.100101}{Atomic-resolution
  chemical mapping using energy-dispersive x-ray spectroscopy}, Phys. Rev. B 81
  (2010) 100101.
\newblock \href {http://dx.doi.org/10.1103/PhysRevB.81.100101}
  {\path{doi:10.1103/PhysRevB.81.100101}}.
\newline\urlprefix\url{https://link.aps.org/doi/10.1103/PhysRevB.81.100101}

\bibitem{Krivanek2012}
T.~C. Lovejoy, Q.~M. Ramasse, M.~Falke, A.~Kaeppel, R.~Terborg, R.~Zan,
  N.~Dellby, O.~L. Krivanek, \href{http://dx.doi.org/10.1063/1.3701598}{Single
  atom identification by energy dispersive x-ray spectroscopy}, Applied Physics
  Letters 100~(15) (2012) 154101.
\newblock \href {http://arxiv.org/abs/http://dx.doi.org/10.1063/1.3701598}
  {\path{arXiv:http://dx.doi.org/10.1063/1.3701598}}, \href
  {http://dx.doi.org/10.1063/1.3701598} {\path{doi:10.1063/1.3701598}}.
\newline\urlprefix\url{http://dx.doi.org/10.1063/1.3701598}

\bibitem{Hyrum13}
H.~S. Anderson, J.~Ilic-Helms, B.~Rohrer, J.~Wheeler, K.~Larson, Sparse imaging
  for fast electron microscopy, IS\&T/SPIE Electronic Imaging (2013)
  86570C--86570C.

\bibitem{LDSampling}
R.~Ohbuchi, M.~Aono, Quasi-monte carlo rendering with adaptive sampling.

\bibitem{Mueller2011}
K.~Mueller, Selection of optimal views for computed tomography reconstruction
  (Jan.~28 2011).

\bibitem{wang2010variable}
Z.~Wang, G.~R. Arce, Variable density compressed image sampling, Image
  Processing, IEEE Transactions on 19~(1) (2010) 264--270.

\bibitem{seeger2008compressed}
M.~W. Seeger, H.~Nickisch, Compressed sensing and bayesian experimental design,
  in: Proceedings of the 25th international conference on Machine learning,
  ACM, 2008, pp. 912--919.

\bibitem{carson2012communications}
W.~R. Carson, M.~Chen, M.~R.~D. Rodrigues, R.~Calderbank, L.~Carin,
  Communications-inspired projection design with application to compressive
  sensing, SIAM Journal on Imaging Sciences 5~(4) (2012) 1185--1212.

\bibitem{Seeger10}
M.~Seeger, H.~Nickisch, R.~Pohmann, B.~Sch\"{o}lkopf, Optimization of k-space
  trajectories for compressed sensing by bayesian experimental design, Magnetic
  Resonance in Medicine 63~(1) (2010) 116--126.

\bibitem{joost2012dynamic}
K.~Joost~Batenburg, W.~J. Palenstijn, P.~Bal{\'a}zs, J.~Sijbers, Dynamic angle
  selection in binary tomography, Computer Vision and Image Understanding.

\bibitem{Vanlier2012}
J.~Vanlier, C.~A. Tiemann, P.~A.~J. Hilbers, N.~A.~W. van Riel, A bayesian
  approach to targeted experiment design, Bioinformatics 28~(8) (2012)
  1136--1142.

\bibitem{merryman2005adaptive}
T.~E. Merryman, J.~Kovacevic, An adaptive multirate algorithm for acquisition
  of fluorescence microscopy data sets, IEEE Trans. on Image Processing 14~(9)
  (2005) 1246--1253.

\bibitem{godaliyaddaMBDS}
G.~M.~D. Godaliyadda, G.~T. Buzzard, C.~A. Bouman, A model-based framework for
  fast dynamic image sampling, in: proceedings of IEEE International Conference
  on Acoustics Speech and Signal Processing, 2014, pp. 1822--6.

\bibitem{Godaliyadda2}
G.~D. Godaliyadda, D.~Hye~Ye, M.~A. Uchic, M.~A. Groeber, G.~T. Buzzard, C.~A.
  Bouman, A supervised learning approach for dynamic sampling, in: IS\&T
  Imaging, International Society for Optics and Photonics, 2016.

\bibitem{specht1991general}
D.~F. Specht, A general regression neural network, IEEE transactions on neural
  networks 2~(6) (1991) 568--576.

\bibitem{suykens1999least}
J.~A. Suykens, J.~Vandewalle, Least squares support vector machine classifiers,
  Neural processing letters 9~(3) (1999) 293--300.

\bibitem{hinton2006reducing}
G.~E. Hinton, R.~R. Salakhutdinov, Reducing the dimensionality of data with
  neural networks, science 313~(5786) (2006) 504--507.

\bibitem{lecun2015deep}
Y.~LeCun, Y.~Bengio, G.~Hinton, Deep learning, Nature 521~(7553) (2015)
  436--444.

\bibitem{krizhevsky2012imagenet}
A.~Krizhevsky, I.~Sutskever, G.~E. Hinton, Imagenet classification with deep
  convolutional neural networks, in: Advances in neural information processing
  systems, 2012, pp. 1097--1105.

\bibitem{hadsell2006dimensionality}
R.~Hadsell, S.~Chopra, Y.~LeCun, Dimensionality reduction by learning an
  invariant mapping, in: Computer vision and pattern recognition, 2006 IEEE
  computer society conference on, Vol.~2, IEEE, 2006, pp. 1735--1742.

\bibitem{he2016deep}
K.~He, X.~Zhang, S.~Ren, J.~Sun, Deep residual learning for image recognition,
  in: Proceedings of the IEEE Conference on Computer Vision and Pattern
  Recognition, 2016, pp. 770--778.

\bibitem{hua2014computer}
K.-L. Hua, C.-H. Hsu, S.~C. Hidayati, W.-H. Cheng, Y.-J. Chen, Computer-aided
  classification of lung nodules on computed tomography images via deep
  learning technique., OncoTargets and therapy 8 (2014) 2015--2022.

\bibitem{wang2016perspective}
G.~Wang, A perspective on deep imaging, IEEE Access 4 (2016) 8914--8924.

\bibitem{park2015deep}
Y.~Park, M.~Kellis, Deep learning for regulatory genomics, Nat Biotechnol
  33~(8) (2015) 825--6.

\bibitem{zhou2014deep}
J.~Zhou, O.~Troyanskaya, Deep supervised and convolutional generative
  stochastic network for protein secondary structure prediction, in:
  International Conference on Machine Learning, 2014, pp. 745--753.

\bibitem{Godaliyadda3}
G.~D. Godaliyadda, D.~Hye~Ye, M.~A. Uchic, M.~A. Groeber, G.~T. Buzzard, C.~A.
  Bouman, A framework for dynamic image sampling based on supervised learning
  (slads), ARXIV, 2017.

\bibitem{Garth}
N.~M. Scarborough, G.~M. D.~P. Godaliyadda, D.~H. Ye, D.~J. Kissick, S.~Zhang,
  J.~A. Newman, M.~J. Sheedlo, A.~U. Chowdhury, R.~F. Fischetti, C.~Das, G.~T.
  Buzzard, C.~A. Bouman, G.~J. Simpson, Dynamic x-ray diffraction sampling for
  protein crystal positioning, Journal of Synchrotron Radiation 24~(1) (2017)
  188--195.
\newblock \href {http://dx.doi.org/10.1107/S160057751601612X}
  {\path{doi:10.1107/S160057751601612X}}.

\bibitem{abadi2016tensorflow}
M.~Abadi, A.~Agarwal, P.~Barham, E.~Brevdo, Z.~Chen, C.~Citro, G.~S. Corrado,
  A.~Davis, J.~Dean, M.~Devin, et~al., Tensorflow: Large-scale machine learning
  on heterogeneous distributed systems, arXiv preprint arXiv:1603.04467.

\bibitem{rowenhorst2006three}
D.~Rowenhorst, J.~Kuang, K.~Thornton, P.~Voorhees, Three-dimensional analysis
  of particle coarsening in high volume fraction solid--liquid mixtures, Acta
  materialia 54~(8) (2006) 2027--2039.

\end{thebibliography}

\end{document}